\documentclass[11pt, a4paper]{article}

\usepackage[utf8]{inputenc}
\usepackage[T1]{fontenc}

\usepackage{amsmath}
\usepackage{amssymb}
\usepackage{amsfonts}
\usepackage{mathtools}
\usepackage{nicefrac}

\usepackage[a4paper, margin=1in]{geometry}
\usepackage{microtype}

\usepackage{graphicx}
\usepackage{booktabs}
\usepackage{longtable}
\usepackage{multirow}
\usepackage{array}
\usepackage{makecell}
\usepackage{ragged2e}
\usepackage{tabularx}
\usepackage{color}
\usepackage{xcolor}

\usepackage[colorlinks,urlcolor=blue,linkcolor=blue,citecolor=blue]{hyperref}
\usepackage{url}

\usepackage{amsthm}
\theoremstyle{definition}
\newtheorem{definition}{Definition}

\newtheorem{thm}{Theorem}
\newcommand{\thistheoremname}{}
\newtheorem{genericthm}[thm]{\thistheoremname}

\usepackage{authblk}

\usepackage[backend=biber,style=numeric-comp,sorting=none]{biblatex}

\newcolumntype{Y}{>{\RaggedRight\arraybackslash}X}
\newcolumntype{L}{>{\footnotesize\RaggedRight\arraybackslash}X}
\newcolumntype{Z}{>{\footnotesize}c}

\usepackage{seqsplit}

\usepackage{xspace}

\newcommand{\tLTN}{\textsc{tLTN}\xspace}

\newcommand{\sLTN}{\textsc{sLTN}\xspace}

\author[1]{Edjard Mota}
\author[2]{Artur d'Avila Garcez}
\author[1]{Maria Azevedo}
\author[1]{Odalisio Neto}
\affil[1]{Instituto de Computação, Universidade Federal do Amazonas}
\affil[2]{Department of Computer Science, City St George's, University of London}

\graphicspath{{common/figures/}}
\usepackage{xr}
\title{Temporal Logic Tensor Networks:\\
       Foundations of Temporal Reasoning in Neural Networks}

\begin{document}
\maketitle

\begin{abstract}
In dynamic scenes---football matches, traffic streams, surgical
procedures, or collaborative assembly---actions unfold at fundamentally
different time scales: in football, a kick happens in milliseconds, a possession
lasts seconds, and complex events such as a successful pass followed by a goal combine
multiple atomic actions over extended durations. Models trained solely
on visual data (sequences of pixel values) struggle to learn effectively from different time scales because they 
lack a principled way of reasoning across fine-grained instants and
broader intervals. Neither a point-wise nor a purely interval semantics of time suffices: the first fragments events into frame-level
snapshots, lacking an overall event identification; the second treats intervals as
atomic truth-values with no access to the state trajectory in each interval.

We introduce \textbf{Temporal Logic Tensor Networks (tLTN)}, a
differentiable neuro-symbolic framework that addresses this duality between events and states
through a formal, multi-sorted logic in which the two temporal
modalities are structurally interdependent. Extending Interval Logic
Tensor Networks (ILTN), \tLTN{} grounds state predicates as fuzzy
truth-value paths over time points, while events are first-class
entities with learnable trapezoidal fuzzy durations using 
differentiable Allen-style relations. A compositional semantics fuses static, event-structural, and state-dynamic satisfaction into a
single differentiable Logic Tensor Network (LTN) grounding.

The central theoretical contribution of this construction is that the
modalities constrain one another bidirectionally: the interval layer
determines \emph{where} point-wise constraints are evaluated, so that temporal operators like \textit{Until} and \textit{Since} are checked within event-defined spans rather than
across the whole trace; the point-wise layer determines \emph{whether}
interval relations are warranted, so that a relation between two
detected extents can be rejected when the state dynamics of the relevant intervals contradict that relation. Through controlled synthetic examples, we illustrate when learning of each single-modality is likely to fail and how the localization mechanism can reduce frame-level computational cost on long traces. 
\end{abstract}

\section{Introduction}
\label{sec:A:intro}

Video understanding, sports analytics, autonomous driving, surgical
workflow monitoring, and human--robot collaboration all share a common
challenge: reasoning about \emph{dynamic scenes} in which actions
unfold at fundamentally different temporal scales. In football, for example, a kick happens in
milliseconds, a possession lasts seconds, and a complex event such as a 
pass followed by a goal combines multiple atomic actions over extended
durations. Choosing the right temporal granularity is critical to learning the relevant events and actions, yet a video feed alone provides no inherent notion of where an action begins or ends, how events relate causally, or what rules define them. Models
trained solely on visual data struggle to resolve this temporal ambiguity,
precisely because they lack a principled way of reasoning across both
fine-grained instants and broader intervals.

Concretely, the learning model sees pixels---sequences of RGB values changing
over time---but it does not know the rules that govern the domain. The learning model is expected to infer that a tackle was legal because the player made contact
with the ball first, or to decide who kicked the ball out when two
players collided in a duel. The video feed alone, no matter how high the resolution and frame rate is fundamentally insufficient to solve such temporal reasoning tasks, e.g. when there are occlusions.

The core challenge is that actions in dynamic scenes can exhibit very
different temporal durations relevant to the outcome of the reasoning task. A dribble may include multiple touches, a shot at goal is instantaneous, a counter-attack can last thirty seconds. The same heterogeneity appears across domains: a cut-in followed by a hard brake
in traffic, a cauterization in surgery, or a part insertion in
collaborative assembly each combine instantaneous transitions with
durative activities. A learning model must handle all of these
simultaneously---reasoning about activities extended in time while maintaining
the precision to pinpoint exact moments---using only a single visual
stream. 

The key insight of this paper is that much of the required temporal knowledge can
be derived from the \emph{dynamics of the scene itself} and embedded
directly into the model. Spatial geometry of the pitch (or road, or
operating table) helps resolve occlusion. Domain rules define what
constitutes a foul, an offside, or a safe maneuver. Temporal logic semantics distinguish a dribble from isolated touches. By encoding such knowledge
as differentiable non-classical logic constraints, we seek to guide data-driven learning without requiring additional sensors, cameras, or expensive labeling
of actions at every frame.

More fundamentally, we argue that the two temporal modalities (actions and events, a kick or a counter-attack) should not apply uniformly over the entire video, but each where it carries information. A counter-attack may contain kicks but it cannot have a very long duration. Interval relations such as a counter-attack unfolding over time are evaluated once, based on detected event extents, and they scope the point-wise temporal facts, such as a kick happening at time \textit{t}. In this setting, a point-wise ``A Until B'' temporal constraint relating events A and B need only be checked within the interval span that those events define, not over the entire trace (all the frames). The saving 
is not in perception, which still runs on every frame, but in 
frame-level constraint evaluation, which will be confined to the interiors of event intervals. 
This is not logic used for 
post-processing; it is the temporal structure of the domain deciding 
where logical evaluation is spent.

To see how the two modalities depend on each other, consider a single
pass from kick to reception. Until the kick is observed, there is no
event to reason about. The moment that \textit{kick} becomes true, the \textit{interval} network layer fixes the start of the pass's temporal scope, and the point-wise layer is activated to check, frame by frame, whether the ball reaches a
teammate before any opponent gains control. If the reception never
arrives, the point-wise trajectory is what reveals that the pass has
failed---the interval layer only reasons about extents proposed as
candidates, such as \textit{reception}, not about its absence. We return to this example in detail in Section~\ref{subsec:A:composite_grounding}.

Our solution is \textbf{Temporal Logic Tensor Networks (tLTN)}, a
differentiable neuro-symbolic framework extending the Logic Tensor Networks (LTN) \cite{serafini2016} to enable reasoning over dynamic
scenes, unifying point-wise and interval-based temporal reasoning from Interval LTN (ILTN) \cite{badreddine2023interval}. 
\tLTN{} organizes its ontology around abstract sorts---Objects, States,
Events \& Actions, and Intervals---and grounds them through a dual
semantic signature: state predicates map to fuzzy truth-value paths
over time steps; events and actions are first-class entities with
learnable trapezoidal fuzzy durations. A compositional \emph{semantic
bridge} combines static, event-structural, and state-dynamic satisfaction
into a single differentiable grounding, and differentiable Allen-style
relations express ordering and overlaps between event
intervals. tLTN enables injecting any such temporal domain knowledge as differentiable
constraints into the loss function of a neural network to shape both learning representation and efficient inference.

The central payoff of the tLTN construction is that the two
modalities can be related seamlessly and interact to improve learning and reasoning. The interval modality provides event identity and a differentiable calculus for qualitative
temporal ordering. The point-wise modality preserves the fine-grained
state evolution that distinguishes a successful action from a near
miss. The interval layer provides the scope for point-wise temporal reasoning, and the point-wise layer constrains which interval relations are admissible. A relation that holds between two detected extents can be rejected when the state 
dynamics (what is true or false point-wise) within the interval contradict the definition of that relation. A single-modality logic would have 
to sacrifice one for the other, which should be possible to represent but inefficiently. \tLTN{} keeps both, at 
the cost of one extra grounding signature and one extra component in 
the semantic bridge, as defined formally in this paper.

The interval LTN machinery is not new: tLTN builds on Interval Logic Tensor 
Networks (ILTN)~\cite{badreddine2023interval}, from which it inherits 
the differentiable Allen-style relations. What ILTN does not provide 
is the point-wise layer; its semantics evaluates properties over 
intervals as atomic truth values, with no access to the state trajectory 
inside the intervals. tLTN retains the ILTN machinery combined with a point-wise reasoning over time\footnote{The point-wise reasoning should be possible in standard LTN by defining special variable \textit{t} alongside the required axioms for time, such as in the case of event calculus with the use of the predicate \textit{happens(x,t)} and adopting linear time. Learning of such point-wise time constraints in practice has been proved difficult in LTN, motivating the adoption of interval LTN.}, introduced as a first-class modality, together with the semantic bridge that binds the two modalities.

The tLTN framework introduced here is applied to football analytics and formalized into a \textbf{Football Interval Temporal
Logic (FITL)} in the companion paper, which defines the FITL axioms with actions and events, illustrates the framework's behavior through a
graphical case study using the \texttt{ballControl} and \texttt{Pass} predicates, and further applies \tLTN{} with the FITL axiomatization to a different case study and benchmark on autonomous driving.


The remainder of the paper is organized as follows: Section 2 contains the required background. Section 3 summarizes neuro-symbolic temporal logic learning. Section 4 introduces the temporal LTN framework. Section 5 discusses the semantics of the dual modality with actions and states. Section 6 concludes the paper and discusses directions for future work.
\section{Background}
\label{sec:A:background}

\subsection{Neuro-Symbolic AI for Video Understanding}

Video understanding remains a core challenge in Artificial
Intelligence because of the semantic gap between low-level perceptual
signals and high-level conceptual reasoning~\cite{tenorio2022spatial}.
While deep learning models excel at frame-level perception---detecting
objects, estimating poses, and classifying atomic actions---they often
fail to capture the compositional, durative, and causal structure of
complex event narratives~\cite{li2021survey, salem2022recurrent}. This
limitation has motivated a surge of interest in neuro-symbolic (NeSy) AI for video understanding by seeking to unify the perceptual power of neural networks
with the capabilities of structured reasoning and formal logic~\cite{marra2024statistical,hitzler2022neuro,garcez2009}.

A prominent NeSy framework is the \textbf{Logic Tensor Network
(LTN)}~\cite{serafini2016logic}, which grounds first-order fuzzy
logic in differentiable vector spaces. In LTN, logical predicates are interpreted by neural models
that map object embeddings to truth values in the interval $[0,1]$. Logical axioms
encoding prior knowledge are incorporated as soft constraints in
the training objective, guiding the model toward representations that
are both data-accurate and logically
consistent~\cite{donadello2017logic}. This approach 
has been argued to offer three key advantages: \textbf{explainability} 
(via traceable inference paths), \textbf{knowledge integration} (through 
symbolic axioms), and \textbf{robustness} (by mitigating spurious correlations) \cite{badreddine2022logic}.

However, standard LTN operates on static relational structures and lack
explicit mechanisms for temporal reasoning---a critical shortcoming for
video analysis, where meaning emerges from the evolution of states and
events over time~\cite{augustine2022visual}. Treating videos as a
sequence of independent frames discards essential context, while
relying solely on implicit modeling of time in neural networks has made the learning task very challenging despite the vast deployment of compute resources and data.

\subsection{Temporal Reasoning in Neuro-Symbolic Systems}

Work within a NeSy framework seeking to address this challenge has proposed to integrate temporal logic into
neuro-symbolic frameworks \cite{NIPS2003_34766559, Perotti2014NeuralNF, borges2011learning}. Temporal logics such as Linear Temporal
Logic (LTL)---with operators $\mathcal{U}$ (Until) and
$\mathcal{S}$ (Since)---provide a compact, formal language for
expressing sequential dependencies, preventing combinatorial explosion
in reasoning over longer sequences~\cite{xu2022don, frigeri2014fuzzy}.
Differentiable variants of temporal logic have been used to enforce
logical consistency in neural outputs over time, allowing systems to
learn not just what happens, but how events
unfold~\cite{DBLP:journals/corr/abs-2404-04007}.

A significant contribution in this direction is \textbf{Interval Logic
Tensor Networks (ILTN)}~\cite{badreddine2023interval}, which extends
LTN to reason over fuzzy temporal intervals rather than discrete time
points. By representing events as trapezoidal fuzzy intervals, ILTN
supports qualitative temporal relations (e.g. \textit{before} and
\textit{overlaps}) inspired by Allen's interval algebra, all within a
fully differentiable framework. This shift from point-wise, state-centric reasoning to
event-centric reasoning promises to capture the nature of
real-world actions with different durations.

Yet, existing practical application results have shown mixed performance. Some applications focus on
point-wise temporal performance losing event identity, while others adopt a pure fuzzy interval semantics, sacrificing compatibility with classical
temporal operators. What is missing is a unified framework that
seamlessly integrates both modalities---enabling reasoning about
instantaneous states and durative events in an integrated learnable architecture.

\subsection{Structural and Temporal Extensions of Logic Tensor Networks}

As neuro-symbolic frameworks evolve to handle more complex data, recent
efforts have extended the fundamental Real Logic of LTNs along distinct dimensions. The two most relevant to our work are
\textbf{Structural Logic Tensor Networks (sLTN)}~\cite{rinaldi2026sltn}, which address
structural organizations of knowledge, and the Interval Logic Tensor
Networks (ILTN), already mentioned, which is extended in this paper to unify point-wise and interval-based temporal reasoning. 

\sLTN{} can be understood within the frame-based tradition: crisp
dimensions with attributes, properties, and operations. This framing is excellent for structural organization and static
scenes, and accounts for graph
topologies and discrete sequence persistence. Technically, \sLTN{}
elevates named tensor axes---time steps, sequence positions, or graph
nodes---to first-class elements in the language by introducing
\emph{structural variables} (e.g. $t$ ranging over indices of a named
dimension) and \emph{structural relations} grounded as fuzzy masks
over index tuples, such as $\operatorname{next}(t, t')$ which is true when $t' = t + 1$. Combined with guarded quantification over pairs such as 
$(t_1, t_2)$ and axis selection (e.g. $x[t][t=0]$ to define a boundary),
\sLTN{} can express persistence, transitions, and boundary conditions
directly at the logic level. In the absence of structural dimensions,
the framework recovers standard LTN semantics as a special case.

For time in particular, this machinery gives \sLTN{} point-wise reasoning 
capabilities: persistence is expressed by universal quantification
guarded by $\operatorname{next}$, transitions by the same mask, and
boundary conditions by axis selection. What \sLTN{} does not provide is
a native notion of an interval with learnable duration, or of interval
relations such as \textit{before}, \textit{meets}, or \textit{overlaps};
those would have to be encoded by the user as additional masks. \sLTN{}
also has no in-built LTL-style operators such as \textit{eventually} or
\textit{always} (the former would be expressed as an existential quantifier over
$t$, and the latter as a universal quantifier). Point-wise temporal reasoning is one of the many possibilities for the very general sLTN, but as far as we know, temporal reasoning has not been sLTN's main focus of application. 

The temporal knowledge addressed by ILTN concerns a different piece of
the puzzle. ILTN extend LTN with a native event
modality: events are first-class entities grounded as trapezoidal fuzzy
intervals, with differentiable Allen-style relations. To our knowledge, \tLTN{} is the first framework in the LTN
family to unify both modalities with a single differentiable semantics: the interval-based event calculus of
ILTN and the LTL-style point-wise layer possible with \sLTN{}'s
structural time dimension.

A reader might ask whether \sLTN{}, with its general axis machinery,
would not subsume the interval-based event reasoning that ILTN and \tLTN{}
provide. Continuous intervals could in principle be encoded as
structural indices with interval-valued attributes, and their relations
as structural constraints, but doing so would require extending
\sLTN{}'s discrete index axes with interval-specific functions and
gradient-preserving interval semantics---a significant extension that
duplicates, rather than reuses, what ILTN already provides. With temporal reasoning as our main focus, we find it easier to extend ILTN here.
The goal is not expressivity in the limit but what the formalism makes
easy to state viz a viz the requirements for efficient neuro-symbolic temporal reasoning and learning, in particular, how naturally the gradients flow through the
corresponding operators and their relationships over time, as specified next. 

\section{Foundations of Differentiable Temporal Reasoning}
\label{sec:A:foundations}

As established in Section~\ref{sec:A:background}, overcoming the
limitations of purely subsymbolic models and frame-based neuro-symbolic
approaches requires an explicit, scalable formalism for temporal
reasoning over actions and events. This section lays the formal groundwork: we first trace the
evolution from discrete-time temporal logic to continuous-time interval
reasoning (Section~\ref{subsec:A:points-to-intervals}), then we recall
the specific computational framework of Interval Logic Tensor Networks
(ILTN) that serves as the event-reasoning substrate for \tLTN{}
(Section~\ref{subsec:A:iltn-formalism}).

\subsection{From Time Points to Time Intervals}
\label{subsec:A:points-to-intervals}

The history of applying formal methods to temporal reasoning is long
and distinguished, with Linear Temporal Logic (LTL) and its variants
serving as the bedrock for system verification and
planning~\cite{pnueli1977temporal}. In the context of neuro-symbolic
AI, these formalisms have been adapted to constrain the outputs of
neural networks over sequential data. Frameworks like Differentiable Temporal Logic (DTL) translate LTL formulae into differentiable loss functions, enabling a network to
learn outputs that are not only accurate with respect to ground-truth
labels but also logically consistent over time~\cite{xu2022don}. For
example, a constraint like
$\texttt{ALWAYS}(\text{action\_A} \rightarrow
\texttt{NEXT}(\text{action\_B}))$ can be enforced, penalizing the
network if it predicts $\text{action\_A}$ without predicting
$\text{action\_B}$ in the subsequent time step. Such approaches
represent a significant advance over static, single-frame definitions,
as they allow reasoning over sequences of indeterminate length. For
instance, a formula using the Until ($\mathcal{U}$) operator can
specify that a condition must hold continuously until another condition
is met---abstracting away the exact number of time steps involved. This
capability is essential in dynamic domains where event durations are
variable and not aligned to fixed temporal grids.

However, this sophisticated LTL-based approach operates on a
fundamentally discrete, \textit{state-centric} view of time, where the
world is modeled as a linear sequence of distinct time points (e.g.
video frames). This can introduce brittleness when dealing with actions
that do not align cleanly with frame boundaries or when analyzing data
from sources with heterogeneous sampling rates.

The core innovation introduced by Interval Logic Tensor Networks
(ILTN)~\cite{badreddine2023interval} is to elevate this representation
from a discrete sequence of states to a truly continuous-time,
\textit{event-centric} model. Instead of reasoning about properties
holding at time points $t, t+1, t+2, \dots$, ILTN reasons about
\textit{fuzzy temporal intervals}---treating events as holistic
entities with learnable, fuzzy durations. In this paradigm, a complex
action is no longer a derived sequence of instantaneous truths but a
first-class object in the ontology, grounded as a trapezoidal fuzzy
interval.

This shift from discrete temporal sequences to continuous fuzzy
intervals provides a more flexible and physically grounded
foundation for reasoning about real-world dynamics. However, as we will
see, a complete framework for dynamic scenes requires
\textit{both} modalities---the state-centric trajectory and the
event-centric interval---unified under a single differentiable
architecture.

\subsection{The Formalism of Interval Logic Tensor Networks (ILTN)}
\label{subsec:A:iltn-formalism}

ILTN provides a concrete computational framework for realizing an event-centric, interval-based temporal logic in differentiable manner. ILTN extends the neuro-symbolic principles of LTN, which ground first-order logic predicates in real-valued vector spaces, by introducing specialized representations for temporal intervals and
their relations~\cite{badreddine2023interval}. The entire system is designed to be end-to-end
differentiable, allowing logical constraints to guide the learning of
neural perception modules via gradient descent~\cite{miro2021end}.

ILTN serves as the core \textit{event-reasoning substrate}
within the more general Temporal Logic Tensor Networks (tLTN)
framework. The formalism of ILTN rests on three core pillars:
trapezoidal fuzzy intervals, differentiable interval relations, and
gradient-preserving smooth operators.

\subsubsection{Trapezoidal Fuzzy Intervals}

To capture the inherent vagueness in the start and end times of
real-world events, ILTN represents the duration of an event not as a
crisp interval $[t_{start}, t_{end}]$ but as a trapezoidal fuzzy
interval. This is a fuzzy set over the real line, defined by four
parameters $a \le b \le c \le d$, which represent the start of the
ramp-up, the start of the full-membership core, the end of the core,
and the end of the ramp-down, respectively. The membership function
$I(x)$ for a point $x$ in time is given by:
\[
I(x) =
\begin{cases}
\frac{x-a}{b-a} & \text{if } x \in (a,b) \\
1               & \text{if } x \in [b, c] \\
\frac{d-x}{d-c} & \text{if } x \in (c, d) \\
0               & \text{otherwise}
\end{cases}
\]

This formulation can model crisp events (when $a=b$ and $c=d$), events
with fuzzy beginnings or endings, and fully fuzzy events. Crucially,
the four parameters $(a, b, c, d)$ can be predicted directly by a
neural network. To simplify the prediction and ensure the ordering
constraint $a \le b \le c \le d$, the neural network is
designed to predict four non-negative values: $(a, b-a, c-b, d-c)$,
which are passed through softplus activations $f(x)=ln(1+e^x)$ to guarantee
positivity.

ILTN also defines semi-infinite intervals to handle operators like
\textit{before} and \textit{after}. For an interval $A=(a,b,c,d)$, the
event $\text{before}(A)$ is represented by the left-infinite interval
$(-\infty, -\infty, a, b)$, and $\text{after}(A)$ is represented by the
right-infinite interval $(c, d, +\infty, +\infty)$.

\subsubsection{Differentiable Allen-style Relations}
\label{subsec:A:diff_allens_rel}

By representing events as fuzzy intervals, ILTN defines a set of fuzzy
temporal relations inspired by Allen's interval algebra. Unlike
classical approaches that would yield a Boolean result, these relations
produce a continuous truth value in $[0,1]$, indicating the degree to
which a relation holds. The main idea is to define these
relations based on the intersections of the fuzzy interval areas.

Let $A$ and $B$ be two trapezoidal fuzzy intervals. The fundamental
relation is containment, defined as:
$$
A \text{ \textbf{in} } B := \frac{|A \cap B|}{|A|}
$$
where $|A|$ denotes the area of the trapezoid $A$, calculated as
$\frac{(c-b)+(d-a)}{2}$, since the height of the trapezoid is equal to 1. The term $|A \cap B|$ represents the area of
the polygon formed by the intersection of the two trapezoids. While the
shape of this intersection can be complex (a polygon with up to six
vertices), its area is analytically derivable by first computing the
coordinates $(x_i,y_i)$ of its vertices and then applying the shoelace formula, where $area = 1/2\lvert \sum_i(x_iy_{i+1} - x_{i+1}y_i) \rvert$.
This analytical derivation is key to the system's overall differentiability,
which we shall use to define the event modality semantics in
Section~\ref{subsec:A:event_layer_semantics}.

\subsubsection{Gradient-Preserving Smooth Operators}

A critical challenge in making a logical framework learnable via
gradient descent is the ``vanishing gradient'' problem. The
trapezoidal membership function $I(x)$ is flat (with a derivative of
zero) in the regions where $I(x)=0$ or $I(x)=1$. The
fundamental relation $A \text{ \textbf{in} } B$ will have zero gradient when the
intervals do not intersect ($A \text{ \textbf{in} } B = 0$) or when
$A$ is fully contained within $B$ ($A \text{ \textbf{in} } B = 1$). If
the gradient is zero, learning simply stops with no signal propagated back through the network to adjust parameters.

ILTN addresses this problem by using distinct functions for
the forward and backward passes of the computation graph. For the
forward pass (calculating the satisfaction value in [0,1]), the exact,
crisp-edged trapezoid function $I(x)$ is used. For the backward pass
(calculating the gradients), a smooth version, $I_{\sim}(x)$, is used.
This smooth version is constructed using the softplus function
$s_{+}(x|\beta) = \frac{1}{\beta}\log(1+e^{\beta x})$, as a
smooth approximation of the ReLU function with non-zero gradients
everywhere.

The smooth membership function $I_{\sim}(x)$ is defined piece-wise
with softplus in a way that ensures its derivative
$\frac{\partial I_{\sim}(x)}{\partial x}$ is non-zero even outside the
interval's support. 
A similar technique is applied to the fundamental relation of containment, defining a smooth version $(A \text{ \textbf{in} } B)_{\sim}$
that provides a learning signal even when the intervals are disjoint or
fully contained~\cite{badreddine2023interval}. 

\section{Temporal Logic Tensor Networks}
\label{sec:A:general-framework}


In this section, we define how tLTN incorporates point-wise temporal logic reasoning over state trajectories into ILTN. Domain-specific languages, such as Football Interval Temporal
Logic, can then be instantiated by populating \tLTN{}'s
abstract ontology with domain-relevant constants, predicates, and
axioms. The full FITL formalization and further domain instantiations and applications are developed in the companion paper.

\subsection{A General Ontology for Dynamic Scene Understanding}
\label{subsec:A:general_ontology}

\tLTN{} grounds a two-sorted logical language (state formulae $\times$
event formulae) in an ontology of four abstract sorts, which can be
instantiated to any dynamic domain, as follows:

\begin{itemize}
    \item \textbf{Objects}: Entities participating in the scene,
          grounded as time-varying feature vectors or static
          embeddings.
    \item \textbf{States}: Time-dependent properties or relations over
          objects, evaluated per time step and yielding a fuzzy
          truth-value trajectory.
    \item \textbf{Events \& Actions}: Durative occurrences represented
          as holistic temporal intervals. We distinguish:
        \begin{itemize}
            \item \textit{Primitive events}: Directly detected from
                  patterns in state trajectories.
            \item \textit{Inferred actions}: Derived via logical
                  composition of primitive events and state constraints.
        \end{itemize}
    \item \textbf{Intervals}: Temporal extents of events, represented
          as trapezoidal fuzzy intervals $(a,b,c,d)$, as defined in
          ILTN.
\end{itemize}

The core innovation is the dual semantic signature:
\begin{itemize}
    \item State predicates map to \textit{truth-value paths}:
          $\langle \ldots \rangle \rightarrow^T$, where $^T$ denotes a
          function from discrete time steps to $[0,1]$.
    \item Event/action predicates map to \textit{fuzzy intervals}:
          $\langle \ldots \rangle \rightarrow I$, where
          $I = (a,b,c,d)$.
\end{itemize}

The combination of the dual semantics enables simultaneous reasoning about instantaneous states
and durative events within a single, end-to-end differentiable
architecture.

We now formalize this ontology through four general definition
templates. In each case, the football-specific constructs used in FITL
(e.g. \texttt{Player}, \texttt{kick}, \texttt{pass}) are provided as illustrative examples.

\begin{definition}[Constant Objects]
\label{def:A:constants}
Let $c$ be a constant symbol in the language. Its signature and
grounding are defined as:
$
c : \langle \rangle \rightarrow \mathbb{R}^{d_c}
$
where $d_c$ is the embedding dimension determined by the perceptual
modality (object detector, tracker, or lookup table). The
grounding $\mathcal{G}(c) \in \mathbb{R}^{d_c}$ is provided by a
domain-specific embedding mechanism.

\noindent\textbf{Example (FITL instantiation):}
\begin{itemize}
    \item $Player : \langle\rangle \rightarrow \mathbb{R}^{d_p}$ ---
          embedding from player detection.
    \item $Ball : \langle\rangle \rightarrow \mathbb{R}^{d_b}$ ---
          embedding from ball detection.
    \item $Team : \langle\rangle \rightarrow \mathbb{R}^{d_t}$ ---
          team identifier.
    \item $Target : \langle\rangle \rightarrow \mathbb{R}^{d_g}$ ---
          goal location.
\end{itemize}
\end{definition}

\begin{definition}[State Relations]
\label{def:A:state_relations}
Let $p$ be a state predicate of arity $n$. Its signature is:
\[
p(o_1, \dots, o_n) : \langle O_1, \dots, O_n \rangle \rightarrow^T
\]
where each $o_i$ is an object of sort $O_i$, and the output is a fuzzy
truth-value path over time. Grounding is achieved via a differentiable
function (a neural network) that processes object embeddings or
spatial coordinates at each time step.

We distinguish two subtypes:
\begin{itemize}
    \item \textbf{Non-spatial predicates} that depend only on object
          embeddings (e.g. team membership).
    \item \textbf{Spatial predicates} that depend on geometric
          relationships, either via explicit coordinates (e.g. $at$)
          or derived proximity measures (e.g. $closeTo$).
\end{itemize}

\noindent\textbf{Example (FITL instantiation):}
\begin{itemize}
    \item $player(P,T) : \langle Player, Team \rangle \rightarrow^T$
          --- non-spatial team membership.
    \item $at(O,x,y) : \langle Object, \mathbb{R}, \mathbb{R} \rangle
          \rightarrow^T$ --- spatial position $pos=(x,y)$ obtained from tracker.
    \item $closeTo(O_1,O_2) : \langle Object, Object \rangle
          \rightarrow^T$ --- spatial proximity via
          $\exp(-\|pos_1 - pos_2\|^2 / \sigma^2)$.
\end{itemize}
\end{definition}

\begin{definition}[Primitive Events]
\label{def:A:primitive_events}
Let $e_p$ be a primitive event predicate of arity $k$ with signature:

\[
e_p(o_1, \dots, o_k) : \langle O_1, \dots, O_k \rangle \rightarrow I
\]
where each $o_i$ is an object of sort $O_i$, and the output is a
trapezoidal fuzzy interval $I = (a,b,c,d)$. Grounding is performed by a
specialized neural module that detects characteristic patterns in
relevant state-relation trajectories (e.g. edges, plateaus or
velocity changes).

\noindent\textbf{Example (FITL instantiation):}
\begin{itemize}
    \item $kick(P,b) : \langle Player, Ball \rangle \rightarrow I$
          --- detected from ball control trajectory.
    \item $gainControl(P,b) : \langle Player, Ball \rangle \rightarrow
          I$ --- positive edge in ball control.
    \item $outOfBounds(b) : \langle Ball \rangle \rightarrow I$ ---
          boundary crossing given $at(b,x,y)$.
\end{itemize}
\end{definition}

\begin{definition}[Inferred Actions/Properties]
\label{def:A:inferred_actions}
Let $p_a$ be an inferred action or durative property of arity $k$. Its
signature is:
\[
p_a(o_1, \dots, o_k) : \langle O_1, \dots, O_k \rangle \rightarrow I
\]
where each $o_i$ is an object of sort $O_i$. Unlike primitive events,
$p_a$ is not directly grounded from perception but is inferred via
logical composition of primitive events and state constraints over a
temporal interval.

\noindent\textbf{Example (FITL instantiation):}
\begin{itemize}
    \item $pass(P,Q,b) : \langle Player, Player, Ball \rangle
          \rightarrow I$
    \item $tackle(R,Q,b) : \langle Player, Player, Ball \rangle
          \rightarrow I$
    \item $goalScored(T,B,G) : \langle Team, Ball, Target \rangle
          \rightarrow I$
\end{itemize}
\end{definition}

\subsection{General Syntax and Grammar of tLTN}
\label{subsec:A:grammar}

\tLTN{} defines a two-sorted first-order language. Any
domain-specific language (e.g. FITL) inherits this grammar and
specializes its vocabulary. The syntax is designed to capture the
complex temporal and logical relationships of dynamic scenarios while
remaining formally defined and computationally tractable. The language separates formulae that describe
properties of state trajectories ($\phi_t$) from those that describe
properties of events and relations between events ($\phi_e$). The key
extension over standard first-order logic is the explicit integration
of LTL-style operators $\mathcal{U}$ for Until and $\mathcal{S}$ for
Since into the event logic.

\begin{definition}[tLTN Language Grammar]
\label{def:A:tLTN_grammar}
The \tLTN{} language consists of:
\begin{enumerate}
    \item \textbf{Terms:}
    \begin{itemize}
        \item \textbf{Object Terms ($t_o$)}: constants $c$ or variables
              $x$.
        \item \textbf{Event Terms ($t_e$)}: $e(t_{o_1},\dots,t_{o_n})$, where $e$ is a primitive event function.
    \end{itemize}

    \item \textbf{State Logic Formulae}
          ($\phi_t$) are evaluated point-wise
          over time and defined recursively in the usual way:
    \[
    \phi_t ::= p(t_{o_1},\dots,t_{o_n}) \mid \neg \phi_t \mid
    \phi_{t_1} \land \phi_{t_2} \mid \phi_{t_1} \lor \phi_{t_2} \mid \phi_{t_1} \to
    \phi_{t_2} \mid \phi_{t_1} \leftrightarrow \phi_{t_2},
    \]
    where $p$ is a state predicate.

    \item \textbf{Event Logic Formulae} ($\phi_e$) are evaluated over
          event intervals, as follows:
    \[
    \begin{aligned}
    \phi_e ::= &\ H(t_e) \mid P(t_e) \mid t_{e_1} \text{ rel } t_{e_2}
               \mid \neg \phi_e \mid \phi_{e_1} \land \phi_{e_2} \mid  \\
               & \forall_e x.\phi_e \mid \exists_e x.\phi_e \mid
               (\phi_{t_1})\,\mathcal{U}\,(\phi_{t_2}) \mid
               (\phi_{t_1})\,\mathcal{S}\,(\phi_{t_2}) \mid \\
               & \bigcirc \phi_t \mid \circledcirc \phi_t,
    \end{aligned}
    \]
    where:
    \begin{itemize}
        \item $H(t_e)$ is the ``happened'' predicate (truth degree of event
              occurrence).
        \item $P(t_e)$ denotes a property of an event (e.g. direction,
              intensity).
        \item \textbf{rel}: Allen-style interval relations (before (\textbf{bf}), overlap (\textbf{ol}), meet (\textbf{mt}), after (\textbf{af}), start (\textbf{st}), during (\textbf{dr}), finish (\textbf{fin}), equals (\textbf{eq})).
        \item $\forall_e, \exists_e$: Quantification over event
              variables.
        \item $\mathcal{U}, \mathcal{S}$: Until and Since operators
              over state formulae, evaluated within event-defined
              scopes.
        \item $\bigcirc \phi_t$: Next-frame operator---asserts that $\phi_t$
              holds at the immediately next time step.
        \item $\circledcirc \phi_t$: Previous-frame operator---asserts that
              $\phi_t$ holds at the immediately prior time step.
    \end{itemize}
\end{enumerate}
\end{definition}

This grammar integrates both temporal modalities while maintaining
compatibility with classical temporal logic and ILTN's interval
semantics. The Until ($\mathcal{U}$) and Since ($\mathcal{S}$) operators
appear within the event logic because they are \emph{scoped} by event
intervals. Their arguments are point-wise state formulae, but their
evaluation domain is restricted to the span defined by the events they
relate, rather than extending over the entire trace. The frame-adjacency
operators ($\bigcirc$, $\circledcirc$) similarly remain part of the
event logic but operate on state formulae, enabling the modeling of
transitions (e.g. loss of control immediately followed by gain of
control in a tackle). This is interpreted via interval relations like
meets (\textbf{mt}) insensitive to discrete frame
boundaries during semantic grounding.

This grammar enables rich cross-modal reasoning. For instance, an event
ordering constraint like $e_1 \textbf{ bf } e_2$ can be conjoined with
a state-level requirement like $(\neg \exists x.\, (q(x)
\,\mathcal{U}\, r(x))$, where the Until operator will be evaluated over the
interval spanning from $e_1$ to $e_2$, as formalized in what follows.

\section{Differentiable Fuzzy Semantics for tLTN}
\label{sec:A:semantics}

The semantics of \tLTN{} define a fully differentiable mapping from
syntactic formulae to fuzzy truth values in $[0,1]$. This grounding connects abstract logical constructs to the continuous outputs
of neural perception modules, enabling end-to-end learning from raw
video data. The semantics are compositional: they begin with primitive
states and events and scale to complex, multi-part actions through
recursive application of fuzzy logical operators.

Having established the general ontological signatures for constants,
state relations, and event predicates in
Section~\ref{subsec:A:general_ontology}, we now detail their grounding
semantics. The distinction between output types, truth-value paths
($\rightarrow^T$) and fuzzy intervals ($\rightarrow I$), is the key of the following discussion.

\subsection{The Semantic Duality: States vs. Events}
\label{subsec:A:semantic_duality}

The ontological
distinction between \textit{states} and
\textit{events} is not merely philosophical but has formal consequences for
logic design~\cite{davis2017logical, borgo2014first}, as follows.

A \textbf{state} is a homogeneous property: if it holds over an
interval, it holds over every sub-interval (e.g. $ballControl(P,b)$ in
football). Its grounding is a \textbf{fuzzy truth-value path}---a
function from discrete time steps (e.g. video frames) to $[0,1]$.

An \textbf{event} (or action) is anti-homogeneous: if it occurs in an
interval, \textit{it does not occur in any proper sub-interval}
(e.g. $pass(P,Q,b)$).\footnote{If a player passes the ball then they cannot pass the ball again until they regain possession.} Its grounding is a \textbf{single trapezoidal
fuzzy interval} $(a,b,c,d)$ representing a holistic temporal extent.

This duality reflects how humans conceptualize dynamic scenes and is
essential for video understanding under real-world variability.
Traditional approaches often
adopt a \textit{uni-modal} semantics---either point-wise (as in LTL) or
purely interval-based (as in Allen's algebra)---which leads to
representational gaps in practice with a corresponding fragmentation of the scene or a loss of state evolution. In the symbolic AI literature, this trade-off leads to the familiar frame problem~\cite{mccarthy1981some, shanahan1997solving}, where a point-wise formalism must state explicitly what persists between 
time steps, while a interval-based approach cannot represent how states may change within an interval. Neither is an error of design; each problem follows from what the respective semantics treats as primitive. Next, we embrace both modalities in a unified tLTN framework.

\subsection{tLTN as a Multi-Modal Logic}
\label{subsec:A:tLTN_multimodal}

Inspired by multi-modal logical systems~\cite{van2006logics, balbiani2006becomes, blackburn2001modal}, \tLTN{} formalizes the dual semantics of states and events as a \textbf{multi-modal logic}: a logic whose vocabulary contains several families of modal operators, each with its own interpretation space. Next, we develop the state-centric and the event-centric modalities, and their semantic bridge. Like in sLTN, the framework admits further modalities. The spatial modality, which contributes $G_{spatial}(A)$
to the grounding of a composite action, is developed in the companion paper. Each modality operates in its own interpretation
space, with dedicated operators and predicates. This structure is key to enabling expressive yet tractable reasoning over
dynamic scenes.

\textbf{Modality 1: the State-Centric Modality
($\mathcal{G}_{\text{state}}$)} is designed for reasoning about
continuous, frame-wise dynamics:
\begin{itemize}
    \item \textbf{Semantic Objects:} Fuzzy truth-value paths (functions
          $\mathbb{N} \to [0,1]$).
    \item \textbf{Interpretation Space:} A Kripke model where
          ``worlds'' are individual time steps (e.g. video frames),
          linked by a linear \textit{next-frame} accessibility
          relation.
    \item \textbf{Operators:} Temporal connectives such as
          $\mathcal{U}$ (Until), $\mathcal{S}$ (Since), $\bigcirc$
          (next), and $\circledcirc$ (previous).
\end{itemize}

\textbf{Modality 2: the Event-Centric Modality
($\mathcal{G}_{\text{event}}$)} is designed for reasoning about durative,
holistic occurrences:
\begin{itemize}
    \item \textbf{Semantic Objects:} Trapezoidal fuzzy intervals
          $(a,b,c,d)$.
    \item \textbf{Interpretation Space:} A universe of temporal
          intervals, where ``accessibility'' is defined by Allen-style
          relations (e.g. \textbf{bf}, \textbf{mt}, \textbf{ol}).
    \item \textbf{Operators:} Interval relational predicates (e.g.
          $e_1 \textbf{ bf } e_2$).
\end{itemize}

\begin{figure}[t]
\centering
\includegraphics[width=0.9\linewidth]{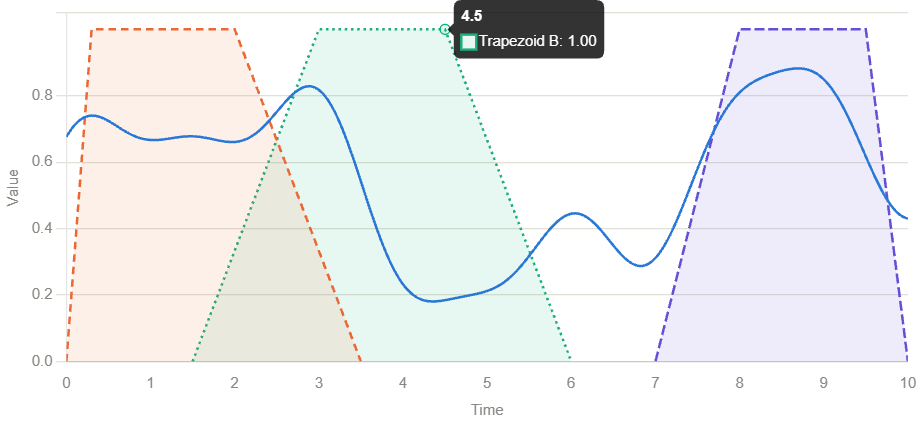}
\caption{The semantic bridge: the solid curve is a state trajectory
(point-wise modality). The dashed trapezoids are fuzzy event extents
(interval modality). At $t=4.5$, the interval for event B (in green) reaches full
membership while the underlying state trajectory is near its minimum, a
configuration that interval-only semantics cannot express, since it
evaluates properties over intervals as atomic truth values, and that
point-wise semantics alone cannot localize, since it has no representation
of the interval in which the state's dip in truth value occurs. Between events B and C (in blue),
point-wise constraint evaluation is localized to the event spans rather
than the full trace, allowing the system to stop reasoning about what should persist at each time step in the absence of an action.
}
\label{fig:semantic-bridge}
\end{figure}

The contribution of \tLTN{} is not the existence of either modality in
isolation, but the \textbf{interaction} between them. Figure~\ref{fig:semantic-bridge} illustrates this interaction. The interval ``layer'' (in reality a neural network) assigns a fuzzy extent
to each event, and the point-wise ``layer'' (an associated neural network) evaluates the state trajectory
\emph{within} that extent. The flow of information is bidirectional: the interval relation determines \emph{where} the
point-wise constraint is evaluated, and the point-wise trajectory determines
\emph{whether} the interval relation is warranted.

Consider the configuration at $t=4.5$ in Figure~\ref{fig:semantic-bridge}.
The interval membership of event B is at its maximum, while the state
trajectory that constitutes B is near its minimum. The goal of tLTN is to express that the constitutive state is failing given the dip in truth value occurring inside the interval of event B. \tLTN{} evaluates the
composite satisfaction as a fuzzy conjunction of the two signals, so that an event whose interval is fully active but whose interior state is failing
receives a reduced truth value. The reduction is not a correction to
either layer; it is what the composition of the two means. 

The same construction yields the computational advantage of the
combination. Because the interval boundaries localize the scope of the
point-wise operators, otherwise subject to the ``frame problem'', an Until constraint relating two events is evaluated
within the span that those events define rather than across the entire trace.
Perception still runs on every frame; what is localized is frame-level
\emph{constraint evaluation}, which is confined to the interiors of
events---precisely where the state trajectory distinguishes a successful
action from a near miss. 

The two modalities are therefore not co-present in \tLTN{} but genuinely
interdependent: without the interval layer, point-wise constraints have no
scope and must be evaluated globally; without the point-wise layer,
interval relations have no interior value and cannot be falsified by changes in the states that they purport to summarize.

\subsection{The Semantic Bridge: Integrating Modalities}
\label{subsec:A:semantic_bridge_tLTN}

What elevates \tLTN{} beyond a mere combination of tools is its
\textbf{compositional semantic bridge}, which formally unifies the two
modalities. The satisfaction of a composite action $A$---such as a pass
or tackle---is computed by a differentiable aggregation function
$\mathcal{T}$ (typically a t-norm) that fuses evidence from both
layers:

\[
\mathcal{G}(A) = \mathcal{T}\big(\mathcal{G}_{\text{static}}(A),\,
\mathcal{G}_{\text{event}}(A),\, \mathcal{G}_{\text{state}}(A)\big)
\]

This bridge ensures that symbolic temporal inferences are jointly constrained by
(1) the qualitative temporal structure of events (e.g. ordering,
overlap) and (2) the continuous evolution of underlying states (e.g.
ball possession, proximity).

\noindent\textbf{Example (FITL instantiation):}
In football analysis, the action $pass(P,Q,b)$ is evaluated by
$\mathcal{G}_{\text{event}}$, the degree to which $kick(P,b)$ \textbf{bf}
$gainControl(Q,b)$, and $\mathcal{G}_{\text{state}}$, the degree to which
$(\neg \exists R.\, gainControl(R,b))\ \mathcal{U}\
gainControl(Q,b)$ holds over the interval between kick and reception.\footnote{Variables $P$ and $Q$ denote players of a team, while variable $R$ denotes a player of the opposing team.} Their conjunction yields a physically grounded satisfaction score for pass validity.


\subsubsection{Grounding of First-Order Terms}
\label{subsec:A:grounding_terms}

Following the principles of Logic Tensor Networks (LTNs) and their
temporal extension ILTN, every term in a \tLTN{}-based language is
grounded in a real-valued vector space. This grounding provides the
link between symbolic terms and continuous perceptual data.

\begin{definition}[Grounding of First-Order Terms in tLTN]
\label{def:A:grounding_terms}
Let $\mathcal{G}$ be the grounding function.
\begin{itemize}
    \item \textbf{Constants:} Every constant symbol $c$ from the
          ontology is grounded onto a tensor (or a multidimensional array denoting a feature vector in
          high-dimensional space): $\mathcal{G}(c) \in \mathbb{R}^d$.
          The dimensionality $d$ and embedding mechanism depend on the
          domain and perceptual modality (e.g. object detector,
          tracker, or static lookup).

    \item \textbf{Variables:} A variable $x$ is grounded not onto a
          single tensor but to a batch of tensors representing a set of
          objects over which quantification occurs:
          $\mathcal{G}(x) \in (\mathbb{R}^d)^b$, where $b$ is the batch
          size (i.e. number of candidate objects in the current
          scene).
\end{itemize}
\end{definition}

\noindent\textbf{Example (FITL instantiation):}
In football video analysis, constants like $Player$, $Ball$, or $Team$
are grounded via embeddings from detection and tracking systems, while
variables range over sets of detected players or teams during
inference.

\subsubsection{Grounding State Relations and Primitive Events}
\label{subsec:A:grounding_state_events}

Atomic formulae form the foundation of \tLTN{}'s semantics. Their
grounding integrates neural perception with symbolic structure,
distinguishing between state-level and event-level representations.

\begin{definition}[Grounding of State Relations]
\label{def:A:grounding_state}
A state relation (predicate) $p$ of arity $n$ is grounded as a
differentiable function $\mathcal{G}(p)$ that maps the embeddings of
$n$ grounded object terms to a \textbf{fuzzy truth-value path}. Let
$o_1, \ldots, o_n$ denote object terms (not time points; the time step
at which the predicate is evaluated is denoted by the subscript $t$).
For grounded terms $\mathcal{G}(o_1), \ldots, \mathcal{G}(o_n)$, the
truth value of $p(o_1, \ldots, o_n)$ at time step $t$ is:
\[
\mathcal{G}_t(p(o_1, \ldots, o_n)) =
\mathcal{G}(p)(\mathcal{G}(o_1), \ldots, \mathcal{G}(o_n))_t \in
[0,1].
\]
The function $\mathcal{G}(p)$ is typically implemented as a 
Transformer or other sequential model that processes object embeddings
over time to produce a trajectory of truth values.
\end{definition}

\begin{definition}[Grounding of Primitive Events]
\label{def:A:grounding_events}
A primitive event predicate $e$ of arity $n$ is grounded as a
specialized neural module $\mathcal{G}(e)$ that maps relevant
\textbf{state-relation truth-value paths} to a single trapezoidal fuzzy
interval $I = (a,b,c,d)$. Let $o_1, \ldots, o_n$ denote the object
terms of the event. Formally:
\[
\mathcal{G}(e(o_1, \ldots, o_n)) =
\mathcal{G}(e)\big(\mathcal{G}(p_1), \ldots, \mathcal{G}(p_m)\big)
\rightarrow I,
\]
where each $\mathcal{G}(p_i)$ is a state trajectory (e.g. proximity,
control, velocity) characterizing the event's perceptual signature.
\end{definition}

This design reflects a key methodological principle of \tLTN{}:
\textbf{primitive events are not detected from raw sensory streams
alone, but from semantically grounded patterns in state trajectories}.
This ensures that event grounding, and the learning of their different time ranges, remains aligned with
domain knowledge and logical constraints.

\noindent\textbf{Example (FITL instantiation):}
In football, the primitive event $kick(P,b)$ is grounded not from pixel
sequences directly, but from characteristic patterns in the
$ballControl(P,b)$ and $at(b,x,y)$ trajectories---such as a sharp drop
in ball control coinciding with a rapid increase in ball velocity.
Similarly, $gainControl(P,b)$ is detected as a rising edge in the
$ballControl$ truth value path. This distinguishes a knowledge-driven 
event detection from the statistics of raw pixel values.

\subsubsection{The Event Modality: Differentiable Semantics for Interval Relations}
\label{subsec:A:event_layer_semantics}

With events grounded as trapezoidal fuzzy intervals, the \textit{Event
Modality} of \tLTN{} requires a fully differentiable calculus for
reasoning about their qualitative temporal relationships. Allen's interval algebra~\cite{allen1983maintaining} defines eight fuzzy temporal relations 
that yield continuous truth values in
$[0,1]$, indicating the degree to which a relation holds between two
events. Following ILTN, these relations 
are defined via the area of intersection of fuzzy trapezoids. The 
containment relation from Section~\ref{subsec:A:diff_allens_rel} serves 
as the primitive building block for the complete set, compositionally defined as follows:

\[
\begin{aligned}
(equal) \quad A \text{ \textbf{eq} } B &:= (A \text{ \textbf{in} } B) \wedge
(B \text{ \textbf{in} } A) \\
(before) \quad A \text{ \textbf{bf} } B &:= A \text{ \textbf{in} } \text{before}(B) \\
(after) \quad A \text{ \textbf{af} } B &:= B \text{ \textbf{in} } \text{after}(A) \\
(meet) \quad A \text{ \textbf{mt} } B &:= \text{end}(A) \text{ \textbf{eq} }
\text{start}(B) \\
(start) \quad A \text{ \textbf{st} } B &:= (\text{start}(A) \text{ \textbf{eq} }
\text{start}(B)) \wedge (\text{end}(A) \text{ \textbf{bf} }
\text{end}(B)) \\
(during) \quad A \text{ \textbf{dr} } B &:= (\text{start}(A) \text{ \textbf{af} }
\text{start}(B)) \wedge (\text{end}(A) \text{ \textbf{bf} }
\text{end}(B)) \\
(finish) \quad A \text{ \textbf{fin} } B &:= (\text{start}(A) \text{ \textbf{af} }
\text{start}(B)) \wedge (\text{end}(A) \text{ \textbf{eq} }
\text{end}(B)) \\
(overlap) \quad A \text{ \textbf{ol} } B &:= (\text{start}(A) \text{ \textbf{bf} }
\text{start}(B)) \wedge (\text{start}(B) \text{ \textbf{bf} }
\text{end}(A)) \wedge \\
& \quad \ (\text{end}(A) \text{ \textbf{bf} }
\text{end}(B))
\end{aligned}
\]

where $\text{before}(A)$ and $\text{after}(A)$ were defined in Section \ref{subsec:A:iltn-formalism}, $\text{start}(A)$ and $\text{end}(A)$ denote small fuzzy
intervals centered at the start and end points of $A$, and $\wedge$
is a fuzzy t-norm (e.g. product or Łukasiewicz). This
compositional design yields a rich, expressive, and fully
differentiable language for describing temporal configurations between events.

\noindent\textbf{Example (FITL instantiation):}
The relation $kick(P,b) \text{ \textbf{bf}
} gainControl(Q,b)$ states that a successful pass begins with a kick and ends
with a teammate receiving the ball. The degree of satisfaction reflects
how well the detected intervals align with this idealized temporal structure.

\subsubsection{Linking States and Events}
\label{subsec:A:composite_grounding}

The most critical component of \tLTN{} is the \textit{semantic
bridge}---a formal mechanism that integrates reasoning across the State
and Event modalities. This bridge enables the evaluation of composite
actions that depend simultaneously on (i) static domain constraints,
(ii) interval-based event structure, and (iii) continuous state
evolution within event-defined scopes.

\paragraph{Interval-scoped grounding:}
Throughout this section we write $\mathcal{G}_{[t_{s}, t_{e}]}(\phi)$ to denote the satisfaction of a state-level formula $\phi$ evaluated over the
discrete time window from $t_{s}$ (start) to $t_{e}$ (end, inclusive), not to be confused with event terms. Formally, this is
the value obtained by applying the compositional semantics of
Definition~\ref{def:A:grounding_state} to the subsequence of truth-value
paths restricted to $\{t_{s}, t_{s+1}, \ldots, t_{e}\}$, with the temporal
operators ($\mathcal{U}$, $\mathcal{S}$, $\Diamond$, $\bigcirc$,
$\circledcirc$) confined to that window. When the endpoints are
themselves derived from event intervals (e.g. $t_s = t_{\text{start}}(E_1)$
and $t_e = t_{\text{end}}(E_2)$), the notation expresses that the
point-wise constraint is evaluated \emph{within the scope} induced by
the surrounding events, rather than across the whole video trace.

\begin{definition}[Grounding of Composite Actions in tLTN]
\label{def:A:composite_grounding}
Let $A$ be a composite action (e.g. $\text{pass}(P, Q, b)$ in
football) defined by:
\begin{itemize}
    \item a set of primitive events $\{E_1, \ldots, E_n\}$,
    \item a set of state-level temporal constraints $\{C_1, \ldots,
          C_m\}$, e.g. LTL formulae,
    \item and optional static predicates (e.g. team membership).
\end{itemize}
The total satisfaction $\mathcal{G}(A) \in [0,1]$ is computed as a
fuzzy conjunction (via a t-norm $\mathcal{T}$) of three components:

\begin{enumerate}
    \item \textbf{Static Constraint Satisfaction
          ($\mathcal{G}_{\text{static}}$):} the degree to which atemporal
          facts hold (e.g. $\text{player}(P, T)$). This is a single
          fuzzy value derived from grounded constants and non-temporal
          predicates.

    \item \textbf{Event Structure Satisfaction
          ($\mathcal{G}_{\text{event}}$):} the degree to which qualitative
          temporal relations among primitive event intervals hold,
          computed using differentiable Allen-style relations. For
          instance, in a pass:
    \[
    \mathcal{G}_{\text{event}} = \mathcal{G}\big(E_{kick}
    \text{ \textbf{bf} } E_{recept}\big),
    \]
    where $E_{kick} = \mathcal{G}(kick(P,b))$ and $E_{recept} =
    \mathcal{G}(gainControl(Q,b))$.

    \item \textbf{State Constraint Satisfaction
          ($\mathcal{G}_{\text{state}}$):} the degree to which temporal
          logic constraints (e.g. Until, Since) are satisfied over the
          \textbf{temporal scope induced by the event intervals}. For
          the \textit{pass(P,Q,b)} axiom from player $P$ to player $Q$:
    $$
    \begin{aligned}
    \mathcal{G}_{state} & \ = \
    \mathcal{G}_{[t_{start}(E_{kick}),\,
    t_{\text{end}}(E_{\text{recept}})]} \\
     & \Big((\neg \exists R.\, \text{gainControl}(R,b)) \ \mathcal{U}
     \ \text{gainControl}(Q,b)\Big) \\
    \end{aligned}
    $$
    where $R$ is a player from the opposing team.
\end{enumerate}

The final satisfaction is:
\[
\mathcal{G}(A) = \mathcal{T}\big(\mathcal{G}_{\text{static}},\,
\mathcal{G}_{\text{event}},\, \mathcal{G}_{\text{state}}\big).
\]
\end{definition}

The semantic bridge ensures that symbolic inferences are jointly
constrained by logical structure, temporal topology, and perceptual
dynamics---enabling physically grounded reasoning across the two
modalities. In FITL, this mechanism allows the system to distinguish a
successful pass from an interception or a misplaced kick, not by
isolated detections, but by holistic multi-modal consistency.

\paragraph{The \texttt{kick}--\texttt{gainControl} example, step by step:}
consider the \texttt{pass} axiom evaluated on a video feed of a football game. In
principle, the Until constraint would be evaluated over the entire feed. The point-wise trajectory would have to be checked at every frame, with $kick$ found to be true by scanning, for example, ball proximity to player and ball velocity. What the interval
layer adds is not expressivity in the limit but scope. Now, the
Until constraint is only evaluated within the span defined by the
events under consideration, and the span itself is a first-class object
with learnable parameters rather than a fixed window chosen by hand.
The bidirectional constraint between the two modalities can be seen in action by tracing the computation over the video feed, as follows:
\begin{enumerate}

  \item \textbf{Before the kick.} The event modality monitors for the
  interval $E_{\text{kick}}$ to become valid. As long as
  $E_{\text{kick}}$ has not been detected, the interval relation
  $E_{\text{kick}}\ \textbf{bf}\ E_{\text{recept}}$ cannot be
  evaluated, and the state constraint
  $(\neg\exists R.\, \text{gainControl}(R,b))\ \mathcal{U}\
  \text{gainControl}(Q,b)$ has no scope. \emph{No frame-level
  monitoring of the state trajectory is required.}

  \item \textbf{When the kick happens.} Once the event modality detects
  $E_{\text{kick}}$, the left endpoint of the interval-scoped grounding
  $\mathcal{G}_{[t_s, t_e]}$ is fixed at
  $t_s = t_{\text{start}}(E_{\text{kick}})$. The point-wise modality is
  now activated. From this frame onward, the truth value of
  $\text{gainControl}(Q,b)$ and of its negation for the opponent players are
  monitored frame by frame, waiting for a right endpoint to close the
  scope.

  \item \textbf{While the pass is in flight.} The interval modality
  maintains a \emph{candidate} for $E_{\text{recept}}$ as soon as any
  player's ball control trajectory truth value starts to rise. The point-wise trajectory
  of $\text{gainControl}(R,b)$ for $R$ in the opponent team is what
  determines whether the candidate is warranted. If any opponent's
  trajectory crosses the threshold before a receiver's trajectory
  does, the Until constraint is violated and the satisfaction of
  $\mathcal{G}_{\text{state}}$ collapses toward zero. The interval
  relation $E_{\text{kick}}\ \textbf{bf}\ E_{\text{recept}}$ may still
  appear as being satisfied on detected extents alone; it is the point-wise
  layer that decides whether the interval relation is
  \emph{warranted}.

  \item \textbf{If the reception never comes.} The Until constraint
  does not force $E_{\text{recept}}$ to exist. If the ball leaves the
  field, or an opponent intercepts it, or the sequence simply ends (e.g. at half-time), the
  point-wise trajectory of $\text{gainControl}(Q,b)$ stays below the
  threshold throughout the interval. The overall satisfaction of \texttt{pass(P,Q,b)} decays smoothly toward zero, indicating a
  failed pass without any explicit rejection step. The absence of the
  second event is detected by the point-wise layer, not by the interval
  layer, because the interval layer only reasons about extents that
  have been proposed as a candidate receiver in this case.
\end{enumerate}

This example makes explicit the bidirectional constraint central to
\tLTN{}. 

\subsection{Compositional Semantics of Logical Formulae}
\label{subsec:A:compositional_semantics}

The truth value of complex formulae in \tLTN{} is computed
compositionally from the truth values of their constituents. For a
state-level formula $\phi$ at time $t$, its grounding
$\mathcal{G}_t(\phi)$ is defined recursively using differentiable fuzzy
logic operators, where $\mathcal{T}$ denotes a t-norm and
$\mathcal{C}$ a t-conorm~\cite{van2022analyzing,
bartl2025differentiable}:

\begin{itemize}
    \item $\mathcal{G}_t(\neg \phi) = 1 - \mathcal{G}_t(\phi)$
    \item $\mathcal{G}_t(\phi \land \psi) =
          \mathcal{T}(\mathcal{G}_t(\phi), \mathcal{G}_t(\psi))$
    \item $\mathcal{G}_t(\phi \lor \psi) =
          \mathcal{C}(\mathcal{G}_t(\phi), \mathcal{G}_t(\psi))$
    \item $\mathcal{G}_t(\phi \rightarrow \psi) =
          \mathcal{C}(1 - \mathcal{G}_t(\phi), \mathcal{G}_t(\psi))$
    \item $\mathcal{G}_t(\forall x: \phi) = \underset{d \in
          D}{\text{Agg}_{\forall}} (\mathcal{G}_t(\phi[d/x]))$
    \item $\mathcal{G}_t(\exists x: \phi) = \underset{d \in
          D}{\text{Agg}_{\exists}} (\mathcal{G}_t(\phi[d/x]))$
\end{itemize}

In these quantification rules, $D$ denotes the \textbf{domain of discourse}, that is, the finite set of all grounded entities (e.g. specific players, the ball, or teams) over which the logical variable $x$ is permitted to range. The notation $\phi[d/x]$ means the syntactic substitution of the variable $x$ with a specific, grounded entity $d \in D$ throughout the formula $\phi$. Consequently, the aggregator $\text{Agg}_{\forall}$ (or $\text{Agg}_{\exists}$) iterates over every instance $d$ within the domain $D$, evaluating the grounded truth value $\mathcal{G}_t(\phi[d/x])$ for each specific substitution, and it combines these values to approximate universal (or existential) quantification. These compositional rules apply uniformly across all \tLTN{}-based theories.

The semantics for temporal operators ($\bigcirc, \circledcirc, \mathcal{U}, \mathcal{S}$) and Allen-style interval relations follow the differentiable formulations
established in ILTN and extended herein, with semantics given by the polygon-area definitions of Section~\ref{subsec:A:diff_allens_rel}).
Together, they provide a fully
differentiable and compositional semantics for the entire \tLTN{}
language family.

\noindent\textbf{Example (FITL instantiation):}
In football analysis, the above rules and aggregations enable the evaluation of tactical
axioms such as team possession and counterattacks, or pass validity as seen in the example above, by composing ground
predicates like $ballControl$ and $player$ using fuzzy logical
connectives and quantifiers, e.g. the first-order logic representation of $\phi \mathcal{U} \psi$ uses quantification: there exists a future moment where $\psi$ holds, and for all moments before it, $\phi$ holds).

Once primitive states and events are grounded, complex inferred actions
are defined by mapping high-level domain concepts into \tLTN{}'s
expressive vocabulary. Original point-wise temporal logic definitions
are systematically lifted into differentiable interval relations and
fuzzy temporal operators, preserving semantics while gaining
graded sensitivity to misalignment over the different time frames at higher levels of abstraction.

\subsection{Formal Semantics of Temporal Logic Operators}
\label{subsec:A:semantics_TempLogic}

Temporal operators in \tLTN{} are interpreted within their respective
modalities---state or event---reflecting the dual ontology of dynamic
scenes.

\paragraph{State Modality Operators ($\mathcal{G}_{\text{state}}$).} They operate on fuzzy truth-value paths and include, mapped to their intended meaning in first-order logic with the quantifiers aggregated in the usual way in LTN:

\begin{itemize}
    \item \textbf{Until ($\mathcal{U}$):} The formula $\phi_1
    \mathcal{U} \phi_2$ requires $\phi_2$ to become true at some point
    $k$, with $\phi_1$ holding at all preceding points. Its
    differentiable LTN semantics is:
    
    \[
    \mathcal{G}_{[t_{\text{start}}, t_{\text{end}}]}(\phi_1
    \mathcal{U} \phi_2) = \max_{k=t_{\text{start}}}^{t_{\text{end}}}
    \left( \mathcal{T} \left( \mathcal{G}_k(\phi_2),
    \min_{j=t_{\text{start}}}^{k-1} \mathcal{G}_j(\phi_1) \right)
    \right)
    \]

    In fuzzy semantics, the $\exists$ and $\forall$ quantifiers in the definition of Until become concrete min-max operators: ``there exists a $k$ witnessing $\phi_2$'' is a \emph{maximum} over $k$ because we are looking for a witnessing point and the best one fixes the value of $k$. Then, ``for all $j < k$, $\phi_1$ holds'' is a \emph{minimum} over the range of $j$ because a single failure anywhere along the sequence must propagate as zero. The t-norm $\mathcal{T}$ conjoins the two groundings. The chosen witness must both satisfy $\phi_2$ at $k$ and be preceded by $\phi_1$ everywhere. In practice, max and min are replaced by smooth approximations so that gradients can flow during training.

    In football, for example:
    
    \[
    \neg \exists R(gainControl(R,b)) \ \mathcal{U} \
    gainControl(Q,b))
    \]
    is evaluated over the interval from the start of a kick event to
    the end of a reception event. 
    yielding a learnable function that defines the optimal
    reception point while penalizing ball interceptions from the opposing team~\cite{xu2022don, frigeri2014fuzzy}.

    \item \textbf{Since ($\mathcal{S}$):} Defined symmetrically for time points in the past, the Since operator follows the same pattern with the two quantifiers swapped. ``There exists $k$ in the past'' maps to a \emph{maximum} over $k$, and ``for all $j$ from $k$ to now'' maps to a \emph{minimum} over $j$:
    \[
    \mathcal{G}_{[t_{\text{start}}, t_{\text{now}}]}(\phi_1
    \mathcal{S} \phi_2) = \max_{k=t_{\text{start}}}^{t_{\text{now}}}
    \left( \mathcal{T} \left( \mathcal{G}_k(\phi_2),
    \min_{j=k+1}^{t_{\text{now}}} \mathcal{G}_j(\phi_1) \right)
    \right)
    \]

%
\end{itemize}

\paragraph{Event Modality Operators ($\mathcal{G}_{\text{event}}$).}
These operate on trapezoidal fuzzy intervals and include the full suite
of Allen-style relations used by ILTN, with semantics defined via intersection areas, as described in
Section~\ref{subsec:A:diff_allens_rel}. These relations enable learning of event ordering, overlaps, and
concurrency---critical for modeling complex interactions in dynamic
domains.

\paragraph{Next and Previous Operators.}
The \textbf{Next} operator ($\bigcirc$) in FITL-based formulations uses
a ``next-frame'' semantics to signify that one event immediately
follows another (e.g. the gain of control by the tackler immediately
follows the loss of control by the dribbler). In \tLTN{}, this
discrete-time adjacency is given a continuous-time semantics using
the meets (\textbf{mt}) interval relation. The formula $E_{loss}
\text{ \textbf{mt} } E_{gain}$ is satisfied to the extent that the
small fuzzy interval representing the end of $E_{loss}$ is equal to the
small fuzzy interval representing the start of $E_{gain}$. This captures the
notion of immediate succession without being rigidly tied to discrete
frame boundaries.

The \textbf{Previous} operator ($\circledcirc$) is defined analogously using the small fuzzy intervals defining the start and end of event intervals. Together with the \textit{Since} operator ($\mathcal{S}$), $\circledcirc$ enables learning of actions that require ``looking back''; for example, \textit{sustained ball control} is defined using multiple
$\circledcirc$ operators, or mapping to the more flexible Since
operator:
$$
(\text{ballControl}(P,b) \land
\text{directionalControl}(P,b)) \ \mathcal{S} \
\text{ballControl}(P,b)
$$
states that the player has been dribbling forward since they initially
gained control. 
This provides a principled way of learning and reasoning about sustained actions of
variable lengths, a significant improvement over checking a fixed
number of past
frames~\cite{borges2007towards}, as evaluated in practice in the companion paper.

\section{Conclusion}
\label{sec:A:conclusion}

This paper introduced \textbf{Temporal Logic Tensor Networks (tLTN)}, a generalized, fully differentiable neuro-symbolic framework that
resolves the historical semantic duality between state-centric and
event-centric temporal reasoning. By treating states as homogeneous
truth-value paths and events as anti-homogeneous trapezoidal fuzzy
intervals, \tLTN{} provides a rigorous multi-modal semantics bridge between the two representations via
a compositional, differentiable satisfaction function. Experimental results to be reported in the companion paper indicate that learning of complex action patterns is
fundamentally more effective when the rules of reasoning about those
patterns are used to induce the network's representational structure.

The companion paper instantiates \tLTN{} into \textbf{Football Interval
Temporal Logic (FITL)}, showing how expert knowledge
ranging from instantaneous kicks to extended, multi-player possessions and tactical plans can be encoded as differentiable rules. It further develops a
graphical case study on the \texttt{ballControl} versus \texttt{pass}
distinction, showing how the dual semantics resolves an
ambiguity that single-modality approaches can struggle with. It also applies the framework to benchmark autonomous driving use cases for neurosymbolic learning and reasoning.

\subsection{Future Horizons: Towards a Unified Neuro-Symbolic Architecture}
\label{subsec:A:future}

The most immediate avenue for future research is the
unification of \tLTN{} with \textbf{Structural Logic Tensor Networks
(sLTN)}. While \tLTN{} masters the temporal duality of dynamic scenes,
\sLTN{} provides a rigorous calculus for topological and
graph-based structural dimensions. By elevating \sLTN{}'s structural
axes to first-class sorts within \tLTN{}'s abstract ontology, we
envision a unified \textit{Spatio-Temporal LTN} capable of
reasoning simultaneously over continuous state dynamics, holistic event intervals, and complex spatial relations.
Concretely, structural dimensions such as \texttt{GraphNode} and \texttt{SequencePos} could be introduced as new abstract sorts 
in \tLTN{}, reusing its existing quantifiers and semantic bridge, as a step toward a unified spatio-temporal LTN.

A second direction is the theoretical analysis of \tLTN{}'s semantic
bridge: characterizing when the multi-modal satisfaction learning is
compositional, when the gradient flow is well-behaved across both
modalities, and under what conditions the interval-based event
reasoning converges during training. These questions are being formalized as a research program dealing with (i) Soundness of Differentiable Interval Semantics, (ii) Compositionality of Tuple Semantics, and (iii) Convergence Guarantees.


\subsection{Theoretical Limitations of Differentiable Fuzzy Logic}
\label{subsec:A:dfl-limitations}

The framework presented in this paper inherits the differentiable
fuzzy logic substrate of LTN and ILTN, and with it a fundamental
theoretical tension: the conflict between strict logical semantics and
numerical optimization. As analyzed in the research work on Differentiable
Fuzzy Logics (DFL) and Logic Tensor
Networks~\cite{van2022analyzing, badreddine2022logic}, standard fuzzy
operators face severe gradient-flow challenges. G\"odel semantics
($\min/\max$) suffer from a \emph{single-passing} pathology, where
gradients propagate to only a single input, causing optimization to be
dominated by outliers and trapping the model in local minima.
Conversely, while product t-norms provide smooth gradients, they
introduce boundary pathologies (vanishing gradients at $0$ or $1$) and
numerical underflow.

LTN resolves these optimization hurdles through its \emph{Stable
Product Real Logic}, which applies linear projections ($\pi_0$,
$\pi_1$) to clamp truth degrees away from exact boundaries. However, as
a mathematical consequence, this stable operator strictly violates
classical t-norm axioms: the identity axiom $T(a,1) = a$ no longer
holds. What works in practice for learning does not correspond to what
the underlying logic prescribes.

This reveals a critical insight: the differentiable logic frameworks that we build on do not truly decouple logical semantic valuation from
computational gradient flow. Instead, they relax or modify the logical
semantics themselves to ensure well-behaved loss gradients. A major
challenge for future research, therefore, is how to explore formalisms that can maintain symbolic truth valuation while being flexible enough to be applicable effectively in practice. The approach adopted by ILTN of decoupling the forward pass in the computational graph from the backpropagation step, by adopting an approximation of the strict logic functions for learning, is worth investigating further in this respect. Another direction is to study novel differentiable operators and network harnesses that might simultaneously satisfy the logic requirements and provide robust, efficient learning dynamics. Addressing this tension will be crucial for the next generation of theoretically sound and highly scalable neuro-symbolic systems.


\printbibliography

\end{document}